# Nonconvex Approach for Sparse and Low-Rank Constrained Models with Dual Momentum

Cho-Ying Wu [a] and Jian-Jiun Ding *[b]

Graduate Institute of Communication Engineering, National Taiwan University,

No. 1, Sec. 4, Roosevelt Rd., Taipei, Taiwan, ROC, 10617, E-mail: r04942049@ntu.edu.tw [a], jjding@ntu.edu.tw [b]

*Abstract*—**In this manuscript, we research on the behaviors of surrogates for the rank function on different image processing problems and their optimization algorithms. We first propose a novel nonconvex rank surrogate on the general rank minimization problem and apply this to the corrupted image completion problem. Then, we propose that nonconvex rank surrogates can be introduced into two well-known sparse and low-rank models: Robust Principal Component Analysis (RPCA) and Low-Rank Representation (LRR). For optimization, we use alternating direction method of multipliers (ADMM) and propose a trick, which is called the dual momentum. We add the difference of the dual variable between the current and the last iteration with a weight. This trick can avoid the local minimum problem and make the algorithm converge to a solution with smaller recovery error in the nonconvex optimization problem. Also, it can boost the convergence when the variable updates too slowly. We also give a severe proof and verify that the proposed algorithms are convergent. Then, several experiments are conducted, including image completion, denoising, and spectral clustering with outlier detection. These experiments show that the proposed methods are effective in image and signal processing applications, and have the best performance compared with state-of-the-art methods.**

## I. INTRODUCTION

Subspace learning is important to pattern recognition and data analysis. It finds the subspaces that fit the statistical model and catches information from high-dimensional data. Face data with illumination variations are known to lie in a low dimensional subspace [1][2]. Videos are high dimensional data but backgrounds themselves lie in a low-dimensional subspace [3]. Principle Component Analysis (PCA) [4] is based on the assumption that all of the data are from the same multivariate Gaussian distribution. In the sense of squared error as error metric, the PCA finds the subspace while preserving the variances of high-dimensional data. However, the presence of missing, incomplete, or outlier data would make the assumption invalid.

To tackle the outlier corruption problem, the Robust Principle Component Analysis (RPCA) [5] was proposed recently:

$$\min_{L,E} rank(L) + \lambda \|E\|_0 \quad \text{s.t. } D = L + E,\tag{1}$$

where $D \in \mathbb{R}^{N_1 \times N_2}$ is the given data matrix, $L$ is the recovered low-rank matrix, $E$ is the recovered sparse matrix with an arbitrary support, and $\lambda$ is a trade-off penalty. The goal of the RPCA is to decompose the data matrix $D$ into a low-rank matrix and a sparse error matrix. Both $L$ and $E$ have the size the same as that of $D$. To make the programming practical, the convex surrogates for $rank(.)$ and $\|.\|_0$ are introduced:

$$\min_{L,E} \|L\|_* + \lambda \|E\|_1 \quad \text{s.t. } D = L + E,\tag{2}$$

where $\|.\|_*$ denotes the nuclear norm and $\|.\|_1$ denotes the $L_1$ norm. The underlying low-rank matrix of the data matrix $D$ can be recovered under the incoherence assumptions [5]. Several algorithms can be applied to solve (2), such as the principle component pursuit (PCP) [5], the accelerated proximal gradient descent (APG) [6], and the inexact augmented Lagrange multiplier (IALM) [7]. Another well-known subspace learning model is the low-rank representation (LRR) [8] as follows:

$$\min_{L,E} \|L\|_* + \lambda \|E\|_\gamma \quad \text{s.t. } D = AL + E,\tag{3}$$

where $A$ is a dictionary matrix spanning the data space and $\gamma$ indicates some regularizers, such as the $L_1$ or $L_{1,2}$ norm. If the singular value decomposition (SVD) of $L^*$ (the optimal solution $L$) is $U\Sigma V^T$, then $UU^T$ (resp. $VV^T$) is uniquely determined by the column space (resp. the row space) of $L^*$ and hence we can refer $UU^T$ (resp. $VV^T$) as the column space (resp. the row space) of $L^*$. From [8] [40], since $UU^T$ can recover the row space of $D$, and the row space of $D$ is determined by the data samples, subspace membership of samples can be decided by $UU^T$. Hence we can use $UU^T$ for clustering.

In these models, the convex surrogates for the $rank(.)$ and the $L_0$ norm $\|.\|_0$ are adopted. However, adopting convex surrogates may obtain suboptimal solutions. The nuclear norm penalizes all of the singular values with the same weights. However, the components with larger singular values are more associated with the matrix structure and thus should be shrunk less. Also, in real-world applications, the underlying matrix of $D$ may not satisfy the incoherence assumptions or the data are corrupted making the optimal solutions differ from the ground truth [9].

Several nonconvex and possibly nonsmooth regularizers with operation on singular values, such as the $L_p$ norm in [10], Capped $L_1$ in [11], Exponential Type Penalty (ETP) in [12], Minimax Concave Penalty (MCP) in [13], and Smoothly

Clipped Absolute Deviation (SCAD) in [14], were proposed and have been shown to be more effective than the nuclear norm surrogate [9] [19] [20]. They are helpful for solving the following generalized rank minimization problem:

$$\min_X f(X) + \lambda \cdot rank(X),$$ (4)

where $f(X)$ is a fidelity term, and $X$ is the data matrix.

The contributions of this manuscript are as follows: (A) First, we propose a novel nonconvex regularizer. It is an even better $rank(.)$ surrogate for the rank minimization problem in (4). (B) From the success of the nonconvex approach on the rank minimization problem, we propose to introduce nonconvex functions as the $rank(.)$ surrogate in the sparse and low-rank constrained models: (i) RPCA and (ii) LRR problems. We use the alternating direction method of multipliers (ADMM) to solve the nonconvex regularized problems. (C) We also propose the dual momentum trick. It adds momentum with a weight on the dual variables of the augmented Lagrangian to get the optimal solution with smaller recovery error. Full theoretical convergence analysis on RPCA and LRR problems with proposed nonconvex approach and proposed momentum trick are derived and organized. (D) Moreover, we prove that the sequences generated from the proposed algorithms are bounded and have the accumulation points satisfying the Karush-Kuhn-Tucker (KKT) condition. We also prove that the convergence rate of the proposed algorithm is the same as that of the convex case. (E) Finally, we conduct extensive experiments, including (i) image completion, (ii) image denoising, and (iii) spectral clustering, to validate the effectiveness of the proposed algorithm in signal and image processing.

**Notations**: We use upper-case letters for matrices or constants, bold-face and lower-case letters for vectors, and normal lower-case letter for scalars. $\|.\|_*$ denotes the nuclear norm, $\|.\|_1$ denotes the $L_1$ norm, $\|.\|_2$ is the $L_2$ norm, $\|.\|_F$ is the Frobenius norm, $<.,.>$ means the inner product, $\partial f$ means the supergradients of a concave function (or subgradients of a convex function), and since supergradients or subgradients at nonsmooth points may not be unique, $\partial f$ is a set. $\sigma_i(A)$ denotes the $i^{th}$ singular value for the matrix $A$, and $A_{ij}$ denotes the $(i, j)^{th}$ element of $A$. $diag(\boldsymbol{w})$ is the vector to diagonal matrix conversion that lays the vector $\boldsymbol{w}$ on the diagonal line.

## II. BACKGROUND AND RELATED WORK

### A. Nuclear norm minimization and optimization

The nuclear norm is the tightest convex relaxation to the $rank(.)$ function [3]. If the nuclear norm is used as the rank surrogate, (4) can be rewritten as:

$$\min_X f(X) + \lambda \|X\|_*.$$ (5)

Here, $f(X)$ is the fidelity term. It is usually the squared error and should satisfy the assumptions as follows.

[**Assumption 1**]

(1) $f : \mathbb{R}^{N_1 \times N_2} \to \mathbb{R}^+$ is differentiable and is $\rho$-Lipschitz, i.e.,

$$\left\| \nabla f(X_1) - \nabla f(X_2) \right\|_F \leq \rho \left\| X_1 - X_2 \right\|_F, \quad \rho > 0, \tag{6}$$

(2) $\inf f(X) > -\infty$.

To solve (5), the proximal gradient descent algorithms are applied [15][16]. Assumption 1 is the sufficient condition for the proximal gradient descent algorithm to be convergent. In the proximal gradient descent algorithms, $f(X)$ at $X^k$ is approximated by a quadratic proximal term:

$$\begin{aligned}
X^{k+1} &= \min_X \nabla f(X^k)^T (X - X^k) + \frac{\mu}{2} \left\| X - X^k \right\|_F^2 + \lambda \left\| X \right\|_* = \min_X \frac{1}{2} \left\| X - \left( X^k - \frac{1}{\mu} \nabla f\left( X^k \right) \right) \right\|_F^2 + \frac{\lambda}{\mu} \left\| X \right\|_* \\
&\triangleq prox_{\frac{\lambda}{\mu} \|\cdot\|_*} \left( X^k - \frac{1}{\mu} \nabla f\left( X^k \right) \right).
\end{aligned} \tag{7}$$

We can use the method of singular value thresholding (SVT) [17] to solve (7). Let

$$P = X^k - \nabla f(X^k) / \mu \tag{8}$$

and the SVD of $P$ is $U \Sigma V^T$. The problem $prox_{\lambda \mu^{-1} \|\cdot\|_*} (P)$ has the closed-form solution as:

$$X^{k+1} = U (\Sigma - (\lambda / \mu) I)_+ V^T, \tag{9}$$

where $(x)_+ = \mathrm{sgn}(x) \cdot \max(|x|, 0)$ is the soft thresholding operator [18].

*B. Nonconvex surrogate for rank minimization*

Since $\| X \|_* = \sum_i \sigma_i(X)$, we use $\sum_i g(\sigma_i(X))$ as the nonconvex surrogate satisfying the Assumption 2.

[**Assumption 2**]

$\hat{g} = \sum_i g(\sigma_i(X))$ is nonconvex and can be nonsmooth. Here, $g(.)$ is concave and nondecreasing with $\sigma_i(X)$. Also, $\partial g$ is finite, i.e., $\inf \partial g > -\infty$ and $\sup \partial g < \infty$.

With the nonconvex form, (4) can be reformulated as:

$$\min_X f(X) + \lambda \sum_i g\left( \sigma_i\left( X \right) \right). \tag{10}$$

TABLE I Some popular nonconvex functions $g(\sigma_i)$ and their supergradients $\partial g(\sigma_i)$ with penalty $\lambda$.

| | $\lambda g(\sigma_i), \ \lambda > 0, \ \sigma_i \geq 0$ | $\lambda \partial g(\sigma_i)$ |
|---|---|---|
| $L_p$ norm [10] | $\lambda \sigma_i^p, \ 0 < p < 1$ | $\lambda p \sigma_i^{p-1}, \ 0 < p < 1$ |
| Capped L$_1$ [11] | $\begin{cases} \lambda \sigma_i & \text{if } \sigma_i < \theta \\ \lambda \theta & \text{if } \sigma_i \geq \theta \end{cases}$ | $\begin{cases} \lambda & \text{if } \sigma_i < \theta \\ [0, \lambda] & \text{if } \sigma_i = \theta \\ 0 & \text{if } \sigma_i \geq \theta \end{cases}$ |
| ETP [12] | $\dfrac{\lambda}{1 - \exp(-\theta)}(1 - \exp(-\theta \sigma_i))$ | $\dfrac{\lambda \theta}{1 - \exp(-\theta)}(\exp(-\theta \sigma_i))$ |
| SCAD [14] | $\begin{cases} \lambda \sigma_i & \text{if } \sigma_i \leq \lambda \\ \dfrac{-\sigma_i^2 + 2\theta \lambda \sigma_i - \lambda^2}{2(\theta - 1)} & \text{if } \lambda < \sigma_i \leq \theta \lambda \\ \lambda^2 (\theta + 1)/2 & \text{if } \sigma_i > \theta \lambda \end{cases}$ | $\begin{cases} \lambda & \text{if } \sigma_i \leq \lambda \\ \dfrac{\theta \lambda - \sigma_i}{\theta - 1} & \text{if } \lambda < \sigma_i \leq \theta \lambda \\ 0 & \text{if } \sigma_i > \theta \lambda \end{cases}$ |
| MCP [13] | $\begin{cases} \lambda \sigma_i - \sigma_i^2/2\theta & \text{if } \sigma_i < \theta \lambda \\ \theta \lambda^2/2 & \text{if } \sigma_i \geq \theta \lambda \end{cases}$ | $\begin{cases} \lambda - \sigma_i/\theta & \text{if } \sigma_i < \theta \lambda \\ 0 & \text{if } \sigma_i \geq \theta \lambda \end{cases}$ |

Also, using the proximal gradient descent algorithm, the iterative updates are as follows:

$$X^{k+1} \triangleq prox_{\frac{\lambda}{\mu}\hat{g}}\left(X^k - \nabla f\left(X^k\right)/\mu\right). \tag{11}$$

Then, the generalized singular value thresholding [19] is applied. Given the problem in (11), we use $P$ as in (8). Different from (7), the proximal operator is associated with the nonconvex function $\hat{g}$. Suppose that the SVD of $P$ is $U\Sigma V^T$. Then, (11) has the closed form solution as

$$X^{k+1} = U(\Sigma - (\lambda/\mu)diag(\boldsymbol{w}))_+ V^T \tag{12}$$

where $\boldsymbol{w}$ is a vector whose elements are $w_i = \partial g(\sigma_i(P^k))$ at the $k$th iteration.

Nonconvex regularizers SCAD and MCP and their supergradients are listed in Table I. The concavity of g(.) is needed under the consideration that the larger singular values should be penalize less, and the smaller ones should be penalized more in the generalized singular value thresholding.

Moreover, there are a variety of iterative methods based on the proximal operator, such as the iteratively reweighted nuclear norm (IRNN) [20], the general iterative shrinkage and thresholding (GIST) [21], and the iterative shrinkage thresholding and reweighted algorithm (ISTRA) [9]. Another famous nonconvex regularizer is the truncated nuclear norm that only shrinks the smaller singular values and is usually applied to the matrix completion problem [22][23].

# III. Proposed Piecewise Regularizer

Fig. 1 illustrates the supergradients of the SCAD and MCP. One can see that, for the SCAD and MCP, their common feature is that the supergradients $\partial g(\sigma_i)$ drop linearly from a value, and $0 \in \partial g(\sigma_i)$ when $\sigma_i$ is above a threshold. The difference is that for the SCAD, the value is held at a certain value below a threshold.

We propose a piecewise regularizer as in Fig. 2. In Fig. 2 we adopt three thresholds to construct a piecewise linear function with four segments, $l_1$, $l_2$, $l_3$, and $l_4$ in the supergradient domain. Note that the structures of $\{l_1, l_2\}$ and $\{l_4, l_3\}$ resemble that of the MCP, i.e., a steep segment followed by a gentle segment. Also, compared with the SCAD, the line segment pair $\{l_2, l_4, l_3\}$ of the proposed regularizer resembles the structure of SCAD with the difference that $l_2$ is not held at a certain value. As a whole, Fig. 2(a) can be seen as a combination of the two MCP-like structures $\{l_1, l_2\}$ and $\{l_4, l_3\}$ and an SCAD-like structure $\{l_2, l_4, l_3\}$.

The singular values smaller than $p_1$ are penalized by larger values and those between $p_1$ and $p_2$ are penalized less. Note that, unlike SCAD and MCP, between $p_1$ and $p_2$, we still treat each singular value differently. Larger ones should be penalized less, and smaller ones should be penalized more. We do not adopt a horizontal segment as in SCAD and MCP, but use a gentle slope. The singular values above $p_3$, which are very large and determine the characteristic of the matrix, are without penalization. Singular values between $p_2$ and $p_3$ are penalized by much smaller values with steeper segment, since in this interval singular values are large but still smaller than those above $p_3$. Thus, we penalize the singular values in this interval with much smaller penalty. Also, not to affect those above $p_3$, the line segment in this interval is designed to be narrow and have a steeper slope. Compared to SCAD, MCP, this multilevel structure is a finer and more delicate way of penalization. The supergradients of the proposed regularizer is formulated as:

$$
\begin{cases}
y = 2 - \dfrac{2 - (a_1 + a_2)}{p_1} \sigma_i & \text{if } \sigma_i \leq p_1, \\[2mm]
y = \dfrac{a_1}{p_1 - p_2} \sigma_i + (a_1 + a_2) - \dfrac{p_1 a_1}{p_1 - p_2} & \text{if } p_1 < \sigma_i \leq p_2, \\[2mm]
y = \dfrac{a_2}{p_2 - p_3} \sigma_i + a_2 - \dfrac{p_2 a_2}{p_2 - p_3} & \text{if } p_2 < \sigma_i \leq p_3, \\[2mm]
0 & \text{otherwise.}
\end{cases}
\tag{13}
$$

Although this formulation seems complicated, one only needs few threshold comparisons and one floating point multiplication for each singular value. Since most of the singular values are small or infinitesimal for a low-rank matrix, the thresholding can be done by only the comparison with $p_1$ in the most cases.

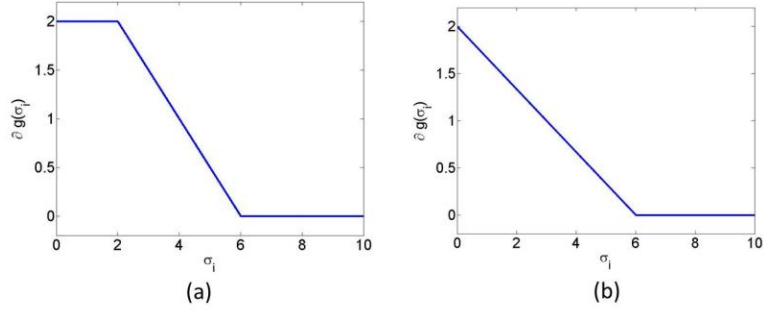

**Fig. 1** Illustrations of supergradients of SCAD and MCP for $\lambda = 2$, $\theta = 3$. (a) SCAD. (b) MCP.

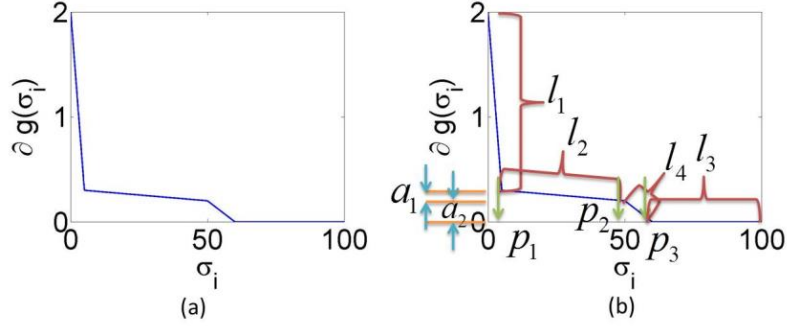

**Fig. 2** (a) The supergradients of the proposed piecewise nonconvex regularizer. (b) With analysis notations.

[**Corollary 1**]:  The proposed piecewise nonconvex surrogate satisfies Assumption 2.

(*Proof*). The concavity of the function g($\sigma_i$) can easily be seen from Fig. 2, its supergradients. Also, since the supergradients are monotonically decreasing from 2, and 0 is the lower bound, the supergradients of the nonconvex surrogate $g$ is finite.

## IV.  NONCONVEX SURROGATE FOR RPCA AND LRR

### A.  Propose to introduce the nonconvex surrogate on RPCA

Both the RPCA and LRR perform sparse and low-rank decomposition. Motivated by the success of the nonconvex surrogate on the rank minimization problem, here we propose to introduce nonconvex surrogates for *rank*(.) in the RPCA and LRR problems. We still use $g$ as the nonconvex regularizer.

The general RPCA problem (1) can be reformulated as:

$$\min_{L,E} \sum_i g(\sigma_i(L)) + \lambda \|E\|_1 \quad \text{s.t. } D = L + E , \tag{14}$$

where we retain the convex surrogate $L_1$-norm for the sparse part and use the nonconvex surrogate $\hat{g}$ to denote the first term in (14) for the low-rank part.

ADMM is a framework for the general programming problem [24][25][26]. According to the ADMM framework, the augmented Lagrangian function can be derived as:

$$\mathcal{L}(L, E, Y, \mu) = \hat{g}(L) + \lambda \|E\|_1 + \langle Y, D - L - E \rangle + \mu \|D - L - E\|_F^2 / 2 , \qquad (15)$$

where $Y$ is the Lagrange multiplier, $\mu > 0$ is a scalar. Subproblems of (15) that update $L$ and $E$ iteratively are:

$$L^{k+1} = \min_L \hat{g}(L) + \frac{\mu^k}{2} \|L - (D - E^k - Y^k / \mu^k)\|_F^2 , \qquad (16)$$

$$E^{k+1} = \min_E \lambda \|E\|_1 + \frac{\mu^k}{2} \|E - (D - L^{k+1} - Y^k / \mu^k)\|_F^2 , \qquad (17)$$

and $\mu^{k+1} = \kappa \mu^k$ with $\kappa > 0$ is the $\mu$-update. The Lagrange multiplier update is:

$$Y^{k+1} = Y^k + \mu^k (D - L^{k+1} - E^{k+1}). \qquad (18)$$

Eq. (16) is in the form of (11), so we can apply (12) to get the closed-form solution for each iteration. Problem (17) is the $L_1$-regularized problem. Therefore, one can use the soft-thresholding operator to get the closed-form solution:

$$E_{ij} = \text{sign}(T_{ij}) \cdot \max \left( |T_{ij}| - \lambda / \mu^k, 0 \right) , \qquad (19)$$

where $T = D - L^{k+1} - (\mu^k)^{-1} Y^k$. The whole algorithm is summarized in Algorithm 1.

---

**Algorithm 1** ADMM for solving RPCA with the nonconvex *rank*(.) surrogate

---

**Input** : Data matrix $D \in \mathbb{R}^{N_1 \times N_2}$ and $\lambda > 0$ .

1) Initialization: $L^0 = 0, E^0 = 0, Y^0 = 0, \mu^0 > 0, \kappa > 1, k = 0$ .

2)    **While** not converge:

3)       Update $L^{k+1}$ by (16).

4)       Update $E^{k+1}$ by (17).

5)       Update $Y^{k+1}$ by (18).

6)       $\mu^{k+1} = \kappa \mu^k$ .

7)       $k = k+1$ .

8)    **Until** converge

**Output**: $(L^k, E^k)$

---

### B. *Propose to introduce the nonconvex surrogate on LRR*

Next, we propose to apply the nonconvex surrogate for *rank*(.) on the LRR problem. The general form of LRR in (3) can be reformulated as:

$$\min_{L,E} \sum_i g(\sigma_i(L)) + \lambda \|E\|_\gamma \quad \text{s.t. } D = AL + E , \tag{20}$$

where $\gamma = 1$ is for random corruptions error, and $\gamma = \{1, 2\}$ is for sample-specific corruption (Here, we adopted the terminology in [8]).

Then, two Lagrangian multipliers, $Y_1$ and $Y_2$, are used instead of $Y$ solely. By introducing another block $Z$ and add "$Z = L$" as another constraint, we can apply the ADMM framework again to write the down the augmented Lagrangian function.

$$\mathcal{L}(L,E,Z,Y_1,Y_2,\mu) = \hat{g}(L) + \lambda \|E\|_\gamma + \langle Y_1, D - AZ - E \rangle + \langle Y_2, Z - L \rangle + \frac{\mu}{2} \|D - AZ - E\|_F^2 + \|Z - L\|_F^2 . \tag{21}$$

The subproblems for updating $L$ can be derived as:

$$L^{k+1} = \min_L \hat{g}(L) + \mu \|L - (Z^k + Y_2^k / \mu^k)\|_F^2 / 2. \tag{22}$$

Again, this is in the form of (11). Therefore, one can use (12) to obtain the closed form solution of (22). Moreover,

$$Z^{k+1} = (I + A^T A)^{-1}(A^T(D - E^k) + L^{k+1}(A^T Y_1^k - Y_2^k) / \mu^k), \tag{23}$$

$$E^{k+1} = \min_E \lambda \|E\|_\gamma + \mu^k \|E - (D - AZ^{k+1} + Y_1^k / \mu^k)\|_F^2 / 2. \tag{24}$$

The updates of the Lagrange multipliers are:

$$Y_1^{k+1} = Y_1^k + \mu^k (D - AZ^{k+1} - E^{k+1}) , \tag{25}$$

$$Y_2^{k+1} = Y_2^k + \mu^k (Z^{k+1} - L^{k+1}), \tag{26}$$

and $\mu$ is updated by $\mu^{k+1} = \kappa \mu^k$ with $\kappa > 0$. The whole algorithm is summarized in Algorithm 2.

---

**Algorithm 2** ADMM for solving LRR with nonconvex *rank(.)* surrogate

---

**Input** : Data matrix $D \in \mathbb{R}^{N_1 \times N_2}$ and $\lambda > 0$ .

1)  Initialization: $L^0 = 0, E^0 = Z^0 = 0, \mu^0 > 0, \kappa > 1, k = 0, Y_1 = 0, Y_2 = 0$ .

2)  **While** not converge:

3)      Update $L^{k+1}$ by (22).

4)      Update $Z^{k+1}$ by (23).

5)      Update $E^{k+1}$ by (24).

6)      Update $Y_1^{k+1}$ by (25) and update $Y_2^{k+1}$ by (26).

7)      $\mu^{k+1} = \kappa \mu^k$ .

8)      $k = k+1$.

9)  **Until** converge

**Output**: $(L^k, Z^k, E^k)$

---

*C. Proposed momentum trick on dual variable.*

On the stochastic gradient descent (SGD), momentum trick is usually used to prevent from getting stuck at the local optimum [41]. The SGD updates the variables not only depending on the gradient of the objective function, but also adding a term accounting for the difference of the variables between the current and the last iterations. Similar to the dynamics that a rolling ball on a slope would not stop instantly when getting into a plane, this difference term is thus called the momentum.

Motivated from the SGD, we propose to use the momentum trick on the updates of dual variables (the Lagrange multiplier). We call this trick the *dual momemtum*. In the nonconvex approach, the local optimum may not equal to the global optimum. Thus, we expect that the momentum term can avoid getting stuck at the local optimum and converge to a solution with smaller recovery residual. Another merit is that it can also speed up the convergence of optimization by introducing a thrust especially when update is too slow.

In the nonconvex RPCA, we introduce another variable sequences $\{\hat{Y}^k\}$ and $\{\alpha^k\}$. Then, the updates of the Lagrange multipliers would be:

$$Y^{k+1} = \hat{Y}^k + \mu^k(D - L^{k+1} - E^{k+1}), \qquad \hat{Y}^{k+1} = Y^{k+1} + \frac{\alpha^k - 1}{\alpha^{k+1}}(Y^{k+1} - Y^k). \tag{27}$$

In this formulation, $\{Y^k\}$ serves as a temporary variable to update $\{\hat{Y}^k\}$. $\{\hat{Y}^k\}$ is the Lagrange multiplier under the dual momentum. Then, from (15), the proposed augmented Lagrangian should be:

$$\mathcal{L}(L, E, \hat{Y}, \mu) = \hat{g}(L) + \lambda\|E\|_1 + \langle \hat{Y}, D - L - E \rangle + \mu\|D - L - E\|_F^2 / 2. \tag{28}$$

Subproblems (16) and (17) should be revised by replacing $Y^k$ with $\hat{Y}^k$.

For $\{\alpha^k\}$, we set $\alpha^0 = 1$ and iteratively determine $\{\alpha^k\}$ from:

$$\alpha^{k+1} = \sqrt{1 + 4(\alpha^k)^2} / 2. \tag{29}$$

Note that $\{\alpha^k\}$ is a monotonically increasing sequence. Also, one can see that

$$\frac{\alpha^{k+1}}{\alpha^k} < \frac{\alpha^k}{\alpha^{k-1}} \tag{30}$$

and $(\alpha^k - 1) / \alpha^{k+1}$ is in the range $[0, 1]$. $\alpha^k$ is a convergent sequence.

The nonconvex RPCA with the proposed dual momentum is organized in Algorithm 3.

---

**Algorithm 3** ADMM for nonconvex RPCA with dual momentum

---

**Input** : Data matrix $D \in \mathbb{R}^{N_1 \times N_2}$ and $\lambda > 0$.

1) Initialization: $\qquad L^0 = 0, E^0 = 0, Y^0 = 0, \hat{Y} = 0, \alpha^0 = 1, \mu^0 > 0, \kappa > 1, k = 0.$

2) **While** not converge:

3)       Update $L^{k+1}$ by (16) but replace $Y^k$ by $\hat{Y}^k$.

4)       Update $E^{k+1}$ by (17) but replace $Y^k$ by $\hat{Y}^k$.

5)       Update $\alpha^{k+1}$ by (29).

6)       Update $Y^{k+1}$ and $\hat{Y}^{k+1}$ by (27).

7)       $\mu^{k+1} = \kappa \mu^k$.

8)       $k = k+1$.

9) **Until** converge

**Output**: $(L^k, E^k)$

---

We also apply the dual momentum on the LRR. Since the LRR adopts three blocks and has two dual variables, as in (21), we would adopt variable sequences $\{\hat{Y}_1^k\}$, $\{\hat{Y}_2^k\}$, and $\{\alpha^k\}$. Then, the dual momentum terms are added on both $\{\hat{Y}_1^k\}$ and $\{\hat{Y}_2^k\}$, which leads to the updates similar to that in (25) and (26):

$$Y_1^{k+1} = \hat{Y}_1^k + \mu^k(D - AZ^{k+1} - E^{k+1}), \qquad \hat{Y}_1^{k+1} = Y_1^{k+1} + \frac{\alpha^k - 1}{\alpha^{k+1}}(Y_1^{k+1} - Y_1^k), \tag{31}$$

$$Y_2^{k+1} = \hat{Y}_2^k + \mu^k(Z^{k+1} - L^{k+1}), \qquad \hat{Y}_2^{k+1} = Y_2^{k+1} + \frac{\alpha^k - 1}{\alpha^{k+1}}(Y_2^{k+1} - Y_2^k). \tag{32}$$

As in the RPCA case, we use $\{\hat{Y}_1^k\}$ and $\{\hat{Y}_2^k\}$ to replace $\{Y_1^k\}$ and $\{Y_2^k\}$ as new Lagrange multipliers. Therefore, the objective function should be revised as:

$$\mathcal{L}(L, E, Z, \hat{Y}_1, \hat{Y}_2, \mu) = \hat{g}(L) + \lambda \|E\|_{\gamma} + \langle \hat{Y}_1, D - AZ - E \rangle + \langle \hat{Y}_2, Z - L \rangle + \mu \left( \|D - AZ - E\|_F^2 + \|Z - L\|_F^2 \right) / 2. \tag{33}$$

Subproblems (22)-(24) should also be revised accordingly by replacing $Y_1^k$ and $Y_2^k$ with $\hat{Y}_1^k$ and $\hat{Y}_2^k$, respectively. The nonconvex LRR with the proposed dual momentum is summarized in the Algorithm 4.

---

**Algorithm 4** ADMM for nonconvex LRR with dual momentum

---

**Input** : Data matrix $D \in \mathbb{R}^{N_1 \times N_2}$ and $\lambda > 0$.

1) Initialization: $L^0 = 0, E^0 = Z^0 = 0, \mu^0 > 0, \kappa > 1, k = 0, Y_1 = 0, Y_2 = 0, \alpha^0 = 1, \hat{Y}_1 = 0, \hat{Y}_2 = 0.$.

2)     **While** not converge:

3)       Update $L^{k+1}$ with (22) but replace $Y_2^k$ by $\hat{Y}_2^k$.

4)       Update $Z^{k+1}$ with (23) but replace $Y_1^k, Y_2^k$ by $\hat{Y}_1^k, \hat{Y}_2^k$

5)       Update $E^{k+1}$ with (24) but replace $Y_1^k$ by $\hat{Y}_1^k$

6)       Update $\alpha^{k+1}$ with (29).

7) Update $Y_1^{k+1}, \hat{Y}_1^{k+1}$ with (31), $Y_2^{k+1}, \hat{Y}_2^{k+1}$ with (32).

8) $\mu^{k+1} = \kappa \mu^k$.

9) $k = k+1$.

10) **Until** converge

**Output**: $(L^k, Z^k, E^k)$

---

In the next section, we will give the full convergence analysis on the proposed nonconvex RPCA and nonconvex LRR algorithms with the dual momentum.

## V. CONVERGENCE ANALYSIS

[**Theorem 1**]: Sequences $\{Y^k\}$ and $\{\hat{Y}^k\}$ generated from Algorithm 3 are bounded.

(*Proof*): From (28),

$$\mathcal{L}(L^{k+1}, E^{k+1}, \hat{Y}^k, \mu^k) = \hat{g}(L^{k+1}) + \lambda \|E^{k+1}\|_1 + \langle \hat{Y}^k, D - L^{k+1} - E^{k+1} \rangle + \mu^k \|D - L^{k+1} - E^{k+1}\|_F^2 / 2.$$

From the partial differentiation with respect to $E^{k+1}$:

$$0 \in \partial_E \mathcal{L}(L^{k+1}, E^{k+1}, \hat{Y}^k, \mu^k), \qquad 0 \in \partial_E(\lambda \|E^{k+1}\|_1) - \hat{Y}^k - \mu^k(D - E^{k+1} - L^{k+1}) = \partial_E(\lambda \|E^{k+1}\|_1) - Y^{k+1}. \qquad (34)$$

From Theorem 3 in [27] about the boundness of the subgradients of norm functions and its dual norm and the fact that dual norm of $L_1$ norm is the $L_\infty$ norm, one can see that $\partial_E(\lambda \|E^{k+1}\|_1)$ is bounded, and thus $\{Y^k\}$ is bounded.

From (27), since $0 \le (\alpha^k - 1) / \alpha^{k+1} < 1$ and $\{Y^k\}$ is bounded, we can see that $\{\hat{Y}^k\}$ is also bounded accordingly.

[**Theorem 2**] Sequences $\{L^k\}$ and $\{E^k\}$ generated from Algorithm 3 are bounded with the assumption that

$$\sum_{k=2}^{\infty} \frac{1}{2(\mu^{k-1})^2}\left[(\mu^k + \mu^{k-1})\|M^k\|_F^2 + 2\frac{\alpha^{k-1}-1}{\alpha^k}\mu^{k-1}\|Y^k - Y^{k-1}\|_F^2 - 2\frac{\alpha^{k-1}-1}{\alpha^k}\frac{\alpha^{k-2}-1}{\alpha^{k-1}}\mu^{k-1}\langle Y^{k-1} - Y^{k-2}, Y^k - Y^{k-1}\rangle\right] < \infty \qquad (35)$$

where $\qquad M^k = Y^k - (1 + \frac{\alpha^{k-2}-1}{\alpha^{k-1}})Y^{k-1} + \frac{\alpha^{k-2}-1}{\alpha^{k-1}}Y^{k-2}.$ $\qquad (36)$

(*Proof*): From the iteration and (28), we have

$$\mathcal{L}(L^{k+1}, E^{k+1}, \hat{Y}^k, \mu^k) \le \mathcal{L}(L^{k+1}, E^k, \hat{Y}^k, \mu^k) \le \mathcal{L}(L^k, E^k, \hat{Y}^k, \mu^k)$$

$$= \mathcal{L}(L^k, E^k, \hat{Y}^{k-1}, \mu^{k-1}) + \frac{1}{2}(\mu^k - \mu^{k-1})\|D - L^k - E^k\|_F^2 + \langle \hat{Y}^k - \hat{Y}^{k-1}, D - L^k - E^k \rangle$$

$$= \mathcal{L}(L^k, E^k, \hat{Y}^{k-1}, \mu^{k-1}) + C^k \qquad (37)$$

where $C^k$ (derived from Lemma 1 in the Appendix) is

$$C^k = \frac{1}{2(\mu^{k-1})^2}\left[(\mu^k + \mu^{k-1})\|M^k\|_F^2 + 2\frac{\alpha^{k-1}-1}{\alpha^k}\mu^{k-1}\|Y^k - Y^{k-1}\|_F^2 - 2\frac{\alpha^{k-1}-1}{\alpha^k}\frac{\alpha^{k-2}-1}{\alpha^{k-1}}\mu^{k-1}\langle Y^{k-1} - Y^{k-2}, Y^k - Y^{k-1}\rangle\right] \quad (38)$$

and $M^k$ is defined as in (36).

From (36), after performing the summation from $k = 2$, we can have:

$$\mathcal{L}(L^{k+1}, E^{k+1}, \hat{Y}^k, \mu^k) \leq \mathcal{L}(L^1, E^1, \hat{Y}^0, \mu^0) + \mathcal{L}(L^2, E^2, \hat{Y}^1, \mu^1) + \sum_{m=2}^{k} C^m. \quad (39)$$

Thus, if $\sum_{m=2}^{\infty} C^m < +\infty$, then $\mathcal{L}(L^{k+1}, E^{k+1}, \hat{Y}^k, \mu^k)$ is upper bounded.

The assumption $\sum_{m=2}^{\infty} C^m < +\infty$ is quite reasonable, since from (30) one can see that $0 \leq (\alpha^k - 1)/\alpha^k < 1$ is a convergent sequence and from Theorem 1 the boundness of $\{Y^k\}$ resulting to that $(Y^{k+1}-Y^k)$ would not diverge to the infinity. One can see that the summation of the latter two terms in the bracket may not diverge to the infinity. Also for the same reason, we can see that $\{M^k\}$ may not be a divergent sequence. Hence we can see that the summation in (35) may not diverge to the infinity.

Next, from (27) and (28)

$$\hat{g}(L^{k+1}) + \lambda\|E^{k+1}\|_1 = \mathcal{L}(L^{k+1}, E^{k+1}, \hat{Y}^k, \mu^k) - 2(\mu^k)^{-1}(\|Y^{k+1} - \hat{Y}^k\|_F^2). \quad (40)$$

Note that, from (39) and Theorem 1, the two terms on the right-hand side of (40) are bounded. Therefore, $\hat{g}(L^{k+1}) + \lambda\|E^{k+1}\|_1$ is bounded, and $\{L^k\}$ and $\{E^k\}$ are bounded, accordingly.

One can see that, with the nonconvex surrogate and the dual momentum, our proof is much more complicated compared with [27] of the ordinary RPCA, and if we take $\alpha^k = \alpha^{k+1} = 1$ for any $k$, $C^k$ reduces to the result in [27].

[**Theorem 3**]: Suppose that the sequences $\{L^k\}, \{E^k\}$, $\{Y^k\}$, and $\{\hat{Y}^k\}$ generated from Algorithm 3 have the accumulation point $(L^*, E^*, Y^*, \hat{Y}^*)$. Assume that $\lim_{k\to\infty} \hat{Y}^{k+1} - \hat{Y}^k = 0$, i.e. $\{\hat{Y}^k\}$ is a Cauchy sequence. Then, $(L^*, E^*, Y^*, \hat{Y}^*)$ is the stationary point satisfying Karush–Kuhn–Tucker (KKT) conditions for problem in (28)

(*Proof*): The primal feasibility is easily from $D - L^{k+1} - E^{k+1} = (\mu^k)^{-1}(Y^{k+1} - \hat{Y}^k)$. Since $\mu^{k+1} = \kappa\mu^k$, we have $\lim_{k\to\infty}(\mu^k)^{-1} = 0$. Therefore,

$$\lim_{k\to\infty} D - L^{k+1} - E^{k+1} = 0. \quad (41)$$

The first order KKT constraints for (28) are:

$$0 \in \partial_E(\lambda \| E^* \|_1) - \hat{Y}^*, \tag{42}$$

$$0 \in \partial_L \hat{g}(L^*) - \hat{Y}^*. \tag{43}$$

To show that (42) holds, from (27),

$$\hat{Y}^{k+1} = \hat{Y}^k + \frac{\alpha^k - 1}{\alpha^{k+1}}(Y^{k+1} - Y^k) + \mu^k(D - L^{k+1} - E^{k+1}). \tag{44}$$

From the assumption that $\lim_{k \to \infty} \hat{Y}^{k+1} - \hat{Y}^k = 0$ and (41),

$$\lim_{k \to \infty}(Y^{k+1} - Y^k) = 0.$$

We can see that $\{Y^k\}$ is also a Cauchy sequence. Then, from (27) again, we can get the property as follows:

$$\lim_{k \to \infty} Y^{k+1} = \hat{Y}^{k+1}. \tag{45}$$

Then, (42) is a direct outcome from (34) with (45), when $k \to \infty$.

To show that (43) holds, from (28), and the optimality of $L^{k+1}$:

$$0 \in \partial_L(\hat{g}(L^{k+1})) - \hat{Y}^k - \mu^k(D - E^{k+1} - L^{k+1}) = \partial_L(\hat{g}(L^{k+1})) - \hat{Y}^k - \frac{\alpha^k - 1}{\alpha^{k+1}}(Y^{k+1} - Y^k) - 2\mu^k(D - E^{k+1} - L^{k+1}). \tag{46}$$

Then, from (41) and that $\{Y^k\}$ is a Cauchy sequence, when $k \to \infty$, we have:

$$0 \in \partial_L(\hat{g}(L^*)) - \hat{Y}^*, \tag{47}$$

and thus the proof is completed.

[**Theorem 4**]: Suppose that Algorithm 3 converges to the primal optimal within finite iterations. Then, for the $K^{\text{th}}$ iteration,

$$\min_{k \in [0, K-1]} \| L^{k+1} - L^k \|_F^2 + \| E^{k+1} - E^k \|_F^2 \leq \frac{1}{\eta K} \left[ \mathcal{L}(L^0, E^0) - \mathcal{L}(L^K, E^K) \right], \tag{48}$$

where $\eta > 0$ is a constant and

$$\mathcal{L}(L, E) = \hat{g}(L) + \lambda \| E \|_1 + H(L, E), \qquad \text{where} \quad H(L, E) = \frac{\mu}{2} \| D - L - E + \mu^{-1} \hat{Y} \|_F^2. \tag{49}$$

Assume that the minimal primal residual of the first $k$ steps always occurs at the $k^{th}$ iteration. Then, the convergence rate of Algorithm 3 is linear.

(*Proof*): Consider that a function $L \to H(L, E)$ is Lipschitz with $L_1$ when we fix $E$, and $E \to H(L, E)$ is Lipschitz with $L_2$ when we fix $L$. Applying Lemma 2 in [28], we have:

$$H(L^{k+1}, E^k) + \hat{g}(L^{k+1}) \leq H(L^k, E^k) + \hat{g}(L^k) - \frac{1}{2}(C_1 - L_1) \| L^{k+1} - L^k \|_F^2, \tag{50}$$

$$H(L^{k+1}, E^{k+1}) + \lambda\big(\big\|E^{k+1}\big\|_1\big) \leq H(L^{k+1}, E^k) + \lambda\big(\big\|E^k\big\|_1\big) - \frac{1}{2}(C_2 - L_2)\big\|E^{k+1} - E^k\big\|_F^2. \tag{51}$$

Here, the constants $C_1$ and $C_2$ satisfy $C_1 > L_1$ and $C_2 > L_2$. Adding (50) and (51), we have:

$$\mathcal{L}(L^{k+1}, E^{k+1}) + H(L^{k+1}, E^k) \leq \mathcal{L}(L^k, E^k) + H(L^{k+1}, E^k) - \frac{1}{2}(C_1 - L_1)\big\|L^{k+1} - L^k\big\|_F^2 - \frac{1}{2}(C_2 - L_2)\big\|E^{k+1} - E^k\big\|_F^2, \tag{52}$$

$$\mathcal{L}(L^k, E^k) - \mathcal{L}(L^{k+1}, E^{k+1}) \geq \frac{1}{2}(C_1 - L_1)\big\|L^{k+1} - L^k\big\|_F^2 + \frac{1}{2}(C_2 - L_2)\big\|E^{k+1} - E^k\big\|_F^2. \tag{53}$$

If we set $\eta = \min(C_1 - L_1, C_2 - L_2)/2$, then:

$$\mathcal{L}(L^k, E^k) - \mathcal{L}(L^{k+1}, E^{k+1}) \geq \eta\Big[\big\|L^{k+1} - L^k\big\|_F^2 + \big\|E^{k+1} - E^k\big\|_F^2\Big]. \tag{54}$$

Last, we use the fact that,

$$\min_{k \in [0, K-1]} \big\|L^{k+1} - L^k\big\|_F^2 + \big\|E^{k+1} - E^k\big\|_F^2 \leq \frac{1}{K}\sum_{k=0}^{K-1} \big\|L^{k+1} - L^k\big\|_F^2 + \big\|E^{k+1} - E^k\big\|_F^2. \tag{55}$$

From (54), we can get:

$$\min_{k \in [0, K-1]} \big\|L^{k+1} - L^k\big\|_F^2 + \big\|E^{k+1} - E^k\big\|_F^2 \leq \frac{1}{\eta K}\Big[\mathcal{L}(L^0, E^0) - \mathcal{L}(L^K, E^K)\Big]. \tag{56}$$

As $K \to \infty$, sequences $\{L^K\}$ and $\{E^K\}$ converge to the limit point $\{L^*\}$ and $\{E^*\}$ proved in Theorem 3, (56) can be reformulated:

$$\min_{k \in [0, K-1]} \big\|L^{k+1} - L^k\big\|_F^2 + \big\|E^{k+1} - E^k\big\|_F^2 \leq \frac{1}{K}\Big[\mathcal{L}(L^0, E^0) - \mathcal{L}(L^*, E^*)\Big] = O(1/K). \tag{57}$$

To date, the proof of the convergence rate of ADMM relies on the convexity of the objective function [42][43][44]. To prove the convergence rate of the nonconvex approach, in (57), we prove that the minimum of the sum of the primal residuals converges with a linear rate. Furthermore, we assume that the minimal primal residual of the first $k$ step always occurs at the $k^{th}$ iteration. This assumption is reasonable. Empirically, the primal residual of ADMM always monotonically decreases considering the stability of the ADMM. Under the assumption, for the $K^{th}$ iteration, (57) can be revised as:

$$\min_{k \in [0, K-1]} \big\|L^{k+1} - L^k\big\|_F^2 + \big\|E^{k+1} - E^k\big\|_F^2 = \big\|L^K - L^{K-1}\big\|_F^2 + \big\|E^K - E^{K-1}\big\|_F^2 \leq \frac{1}{\eta K}\Big[\mathcal{L}(L^0, E^0) - \mathcal{L}(L^*, E^*)\Big] = O(1/K). \tag{58}$$

Thus, we prove the linear convergence of the Algorithm 3.

Then, we show the converge properties of LRR. The proofs of the convergence properties of LRR are similar to those of RPCA in Theorems 1-4.

[**Theorem 5**]:  Sequences $\{L^k\},\{E^k\},\{Z^k\},\{Y_1^k\},\{Y_2^k\},\{\hat{Y}_1^k\}$,  and $\{\hat{Y}_2^k\}$  generated from Algorithm 4 are bounded.

(*Proof*): Note that, in Algorithm 4, (24) in Algorithm 2 should be modified as

$$E^{k+1} = \min_E \lambda \|E\|_\gamma + \mu^k \left\| E - (D - AZ^{k+1} + \hat{Y}_1^k / \mu^k) \right\|_F^2 / 2.$$

From the optimality of $E^{k+1}$,

$$0 \in \partial_E \ \lambda \|E^{k+1}\|_\gamma + \mu^k \left\| E^{k+1} - (D - AZ^{k+1} + \hat{Y}_1^k / \mu^k) \right\|_F^2 / 2 \ ,$$

$$0 \in \partial_E (\lambda \|E^{k+1}\|_\gamma) - \hat{Y}_1^k - \mu^k (D - AZ^{k+1} - E^{k+1}) = \partial_E (\lambda \|E^{k+1}\|_\gamma) - Y_1^{k+1}. \tag{59}$$

Similarly, from (22) and the optimality of $L^{k+1}$,  $0 \in \partial_L \hat{g}(L^{k+1}) - \hat{Y}_2^k - \mu^k (Z^{k+1} - L^{k+1})$ . From (32),

$$0 \in \partial_L \hat{g}(L^{k+1}) - Y_2^{k+1}. \tag{60}$$

Then, following the process similar to that in Theorem 1, one can prove that $\{Y_1^k\}$, $\{Y_2^k\}$, $\{\hat{Y}_1^k\}$, and $\{\hat{Y}_2^k\}$  are bounded. Furthermore, using the procedure similar to that in Theorem 2, one can show that $\{L^k\}$, $\{E^k\}$, and $\{Z^k\}$ updated by line (4) in Algorithm 4 are also bounded.

[**Theorem 6**]:  Suppose that sequences $\{L^k\},\{E^k\},\{Z^k\},\{Y_1^k\}$, $\{Y_2^k\}$, $\{\hat{Y}_1^k\}$, and $\{\hat{Y}_2^k\}$ generated from Algorithm 4 have the accumulation point denoted by $(L^*, E^*, Z^*, Y_1^*, Y_2^*, \hat{Y}_1^*, \hat{Y}_2^*)$. Assume that $\lim_{k\to\infty} \hat{Y}_1^{k+1} - \hat{Y}_1^k = 0$ and $\lim_{k\to\infty} \hat{Y}_2^{k+1} - \hat{Y}_2^k = 0$, i.e. $\{\hat{Y}_1^k\}$ and $\{\hat{Y}_2^k\}$ are Cauchy sequences. Then, $(L^*, E^*, Z^*, Y_1^*, \hat{Y}_1^*, Y_2^*, \hat{Y}_2^*)$ is the stationary point satisfying KKT conditions for the problem in (33).

(*Proof*): From (31) and (32), the primal feasibility is:

$$D - AZ^{k+1} - E^{k+1} = (\mu^k)^{-1} (Y^{k+1} - \hat{Y}^k), \tag{61}$$

$$Z^{k+1} - L^{k+1} = \mu^{k-1} Y_2^{k+1} - \hat{Y}_2^k . \tag{62}$$

The first order KKT constraints follow from the optimality of $L^{k+1}$, $E^{k+1}$, and $Z^{k+1}$ with the assumption that $\lim_{k\to\infty} \hat{Y}_1^{k+1} - \hat{Y}_1^k = 0$ and $\lim_{k\to\infty} \hat{Y}_2^{k+1} - \hat{Y}_2^k = 0$ . Then, from the process similar to that in Theorem 3, one can show that these sequences converge to the stationary point $(L^*, E^*, Z^*, Y_1^*, \hat{Y}_1^*, Y_2^*, \hat{Y}_2^*)$ .

[**Theorem 7**]:  Assume that Algorithm 4 converges to the primal optimal within finite iterations. Then, for the $K_2^{th}$ iteration,

$$\min_{k\in[0,K_2-1]}\left\|L^{k+1}-L^k\right\|_F^2+\left\|E^{k+1}-E^k\right\|_F^2+\left\|Z^{k+1}-Z^k\right\|_F^2\leq\frac{1}{\eta_2 K_2}\left[\mathcal{L}(L^0,E^0,Z^0)-\mathcal{L}(L^{K_2},E^{K_2},Z^{K_2})\right],\tag{63}$$

where $\eta_2>0$ is a constant and

$$\mathcal{L}(L,E,Z)=\hat{g}(L)+\lambda\|E\|_\gamma+H_2(L,E,Z),\tag{64}$$

where $\quad H_2(L,E,Z)=\mu\left\|D-AZ-E+\mu^{-1}\hat{Y}_1\right\|_F^2+\left\|Z-L+\mu^{-1}\hat{Y}_2\right\|_F^2/2.$

Assume that the minimal primal residual of the first $k$ steps always occurs at the $k^{th}$ iteration. Then, the convergence rate of Algorithm 4 is linear.

(*Proof*): Following the process similar to that in Theorem 4, we can use Lemma 2 in [28] again to derive that:

$$\mathcal{L}(L^k,E^k,Z^k)-\mathcal{L}(L^{k+1},E^{k+1},Z^{k+1})\geq\eta_2\left[\left\|L^{k+1}-L^k\right\|_F^2+\left\|E^{k+1}-E^k\right\|_F^2+\left\|Z^{k+1}-Z^k\right\|_F^2\right],\tag{65}$$

where $\eta_2>0$ is a constant. Then, from (65)

$$\begin{aligned}\min_{k\in[0,K_2-1]}\left\|L^{k+1}-L^k\right\|_F^2+\left\|E^{k+1}-E^k\right\|_F^2+\left\|Z^{k+1}-Z^k\right\|_F^2&\leq\frac{1}{K_2}\sum_{k=0}^{K_2-1}\left\|L^{k+1}-L^k\right\|_F^2+\left\|E^{k+1}-E^k\right\|_F^2+\left\|Z^{k+1}-Z^k\right\|_F^2\\&\leq\frac{1}{\eta_2 K_2}\mathcal{L}(L^0,E^0,Z^0)-\mathcal{L}(L^{K_2},E^{K_2},Z^{K_2}).\end{aligned}\tag{66}$$

As $K_2\to\infty$, sequences $\{L^{K_2}\},\{E^{K_2}\}$, and $\{Z^{K_2}\}$ converge to the limit point $\{L^*\},\{E^*\}$, and $\{Z^*\}$ proved in Theorem 6. Then, (66) can be reformulated as:

$$\min_{k\in[0,K_2-1]}\left\|L^{k+1}-L^k\right\|_F^2+\left\|E^{k+1}-E^k\right\|_F^2+\left\|Z^{k+1}-Z^k\right\|_F^2=\frac{1}{\eta_2 K_2}\left[\mathcal{L}(L^0,E^0,Z^0)-\mathcal{L}(L^*,E^*,Z^*)\right].\tag{67}$$

Under the similar assumption as in Theorem 4: the minimal primal residual of the first $k$ step always occurs at the $k^{th}$ iteration, we can revise (67) as:

$$\begin{aligned}\min_{k\in[0,K_2-1]}\left\|L^{k+1}-L^k\right\|_F^2+\left\|E^{k+1}-E^k\right\|_F^2+\left\|Z^{k+1}-Z^k\right\|_F^2&=\left\|L^{K_2}-L^{K_2-1}\right\|_F^2+\left\|E^{K_2}-E^{K_2-1}\right\|_F^2+\left\|Z^{K_2}-Z^{K_2-1}\right\|_F^2\\&\leq\frac{1}{\eta K_2}\left[\mathcal{L}(L^0,E^0,Z^0)-\mathcal{L}(L^*,E^*,Z^*)\right]=O(1/K_2).\end{aligned}\tag{68}$$

Thus, the linear convergence of the Algorithm 4 is proved.

## VI. Experimental Results

In this section, several simulations are conducted to test the performances of the proposed regularizer, the proposed nonconvex RPCA with dual momentum, and the proposed nonconvex LRR with dual momentum. The related codes can be downloaded from [45].

TABLE II

Relative error comparison with the state-of-the-art nonconvex regularizers for low-rank matrix completion.

| $d$ | SCAD | MCP | $L_p$ norm | Capped $L_1$ | ETP | **Proposed** |
|---|---|---|---|---|---|---|
| 10 | 1.50E-13 | 1.41E-13 | 2.05E-08 | 1.32E-12 | 4.16E-11 | **1.03E-13** |
| 12 | 6.96E-12 | 6.57E-12 | 2.59E-08 | 7.69E-12 | 9.76E-11 | **7.44E-12** |
| 14 | 3.14E-10 | 3.06E-10 | 3.30E-08 | 1.08E-10 | 6.57E-10 | **3.19E-10** |
| 16 | 4.29E-09 | 4.18E-09 | 4.98E-08 | 9.41E-09 | 2.39E-09 | **4.85E-09** |
| 18 | 2.76E-08 | 2.65E-08 | 1.01E-07 | 1.66E-07 | 2.14E-08 | **1.64E-08** |
| 20 | 1.95E-07 | 1.87E-07 | 5.01E-07 | 7.55E-06 | 3.11E-07 | **1.64E-07** |
| 22 | 3.51E-06 | 3.30E-06 | 1.20E-06 | 9.35E-05 | 1.66E-06 | **1.57E-06** |
| 24 | 4.90E-05 | 4.79E-05 | 5.04E-06 | 0.000351 | 1.07E-05 | **1.13E-05** |
| 26 | 0.000146 | 0.000139 | 1.86E-05 | 0.001304 | 0.150206 | **2.52E-05** |

TABLE III

Average computational time of the proposed and the state-of-the-art nonconvex regularizers for low-rank matrix completion

| $d$ | SCAD | MCP | $L_p$ norm | Capped $L_1$ | ETP | Proposed |
|---|---|---|---|---|---|---|
| 10 | 1.4563 | 1.4504 | 1.9153 | 1.2743 | 1.1231 | **1.0726** |
| 12 | 1.6110 | 1.5936 | 1.9454 | 1.3030 | **1.3192** | 1.3254 |
| 14 | 1.7893 | 1.7510 | 1.9234 | 1.4308 | 1.4385 | **1.3627** |
| 16 | 1.9395 | 1.9044 | 1.9804 | 1.6191 | 1.3859 | **1.3556** |
| 18 | 2.1453 | 2.1278 | 1.9646 | 1.8202 | **1.4148** | 1.4198 |
| 20 | 2.4068 | 2.3733 | 1.9824 | 2.0131 | 1.4096 | **1.3759** |
| 22 | 2.6227 | 2.6314 | 2.0278 | 2.2507 | **1.4657** | 1.4657 |
| 24 | 2.8861 | 2.8416 | 2.0600 | 2.4778 | 1.5581 | **1.5632** |
| 26 | 3.1576 | 3.0893 | 2.1351 | 2.8222 | 1.7193 | **1.6980** |

### A. Proposed Nonconvex Surrogate

First, the proposed nonconvex regularizer is applied to the low-rank matrix completion problem as follows:

$$\min_X \ \sum_i g(\sigma_i(X)) + \frac{1}{2}\left\| P_\Omega(X) - P_\Omega(O) \right\|_F^2 \ , \tag{69}$$

where $g(\sigma_i(X))$ is the nonconvex surrogate, and $\Omega$ is the support of the matrix, $P_\Omega$ is the linear mapping onto $\Omega$ (keeping the entries in $\Omega$ and set others to zero), and $O$ is the underlying uncorrupted matrix. After projection, we can obtain the corrupted matrix $P_\Omega(O)$. The optimization follows the singular value thresholding in Section II.

For the parameter setting of the proposed regularizer, $(a_1, a_2)$ are much smaller penalty than the largest penalty, i.e. 2. Hence we fix $(a_1, a_2) = (0.1, 0.2)$. For $(p_1, p_2, p_3)$, since we want most singular value thresholdings can be done by comparing to $p_1$ only, the decision of $p_1$ is the most significant. To set $(p_1, p_2, p_3)$, we can roughly investigate the range of the singular values of the examined type of data. Then, we choose $p_1$ to make only the top largest 5% of the singular values are larger than $p_1$, and choose $p_2$ and $p_3$ to make only the top largest 2%, 1% of the singular values are larger than $p_2$ and $p_3$, respectively.

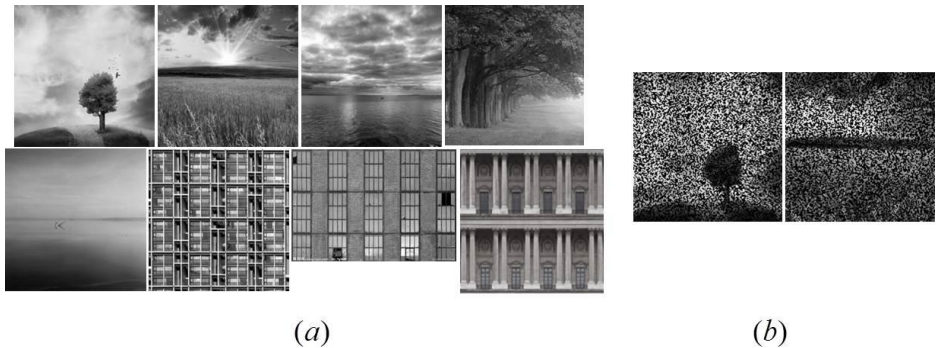

*(a)*                      *(b)*

**Fig. 3** (a) Sample image set (no. 1-4 for the upper, and no. 5-8 for the lower) for corrupted image recovery. (b) Illustrations of corrupted images.

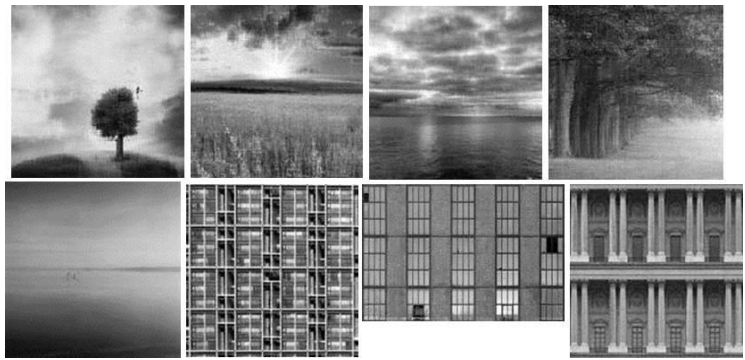

**Fig. 4** The recovered image set with the proposed regularizer.

### A.1 Random matrix synthetic data

In the first experiment, we generate two i.i.d. random matrices $M_1 \in \mathbb{R}^{m \times d}$ and $M_2 \in \mathbb{R}^{d \times n}$ and let $O = M_1 M_2$. We set $m = n = 150$, $d \in [10, 26]$, and $\Omega$ is a set consisting of half of the indexes (randomly chosen) from $m \times n$. Namely, half of the entries are corrupted. We use the partial SVD to compute up to 30 singular values instead of the full SVD to save time.

We first compare the proposed regularizer in Fig. 2 with $(a_1, a_2) = (0.1, 0.2)$ and $(p_1, p_2, p_3) = (5, 50, 60)$ to other regularizers in Table I: SCAD [14] with $\theta = 10$; MCP [13] with $\theta = 10$; $L_p$ norm [10] with $\theta = 0.25$; Capped $L_1$ [11] with $\theta = 70$; ETP [12] with $\theta = 0.1$. The parameters of the compared methods are *tuned to have the best performance.*

We calculate the relative error $= \|X - O\|_F / \|O\|_F$ in Table II and show the computational time in Table III. For each $d$, the experiments are repeated 100 times with different random matrices and different $\Omega$ for evaluation. From Tables II and III, one can see that proposed regularizer attains the smallest relative error within the minimal computation time.

### A.2 Corrupted image recovery

TABLE IV   PSNR comparison with other state-of-the-art methods. ETNNR-WRE and TNNR-WRE are immeasurable for No.7 because they cannot compute nonsquare matrix.

| No. | SCAD | MCP | $L_p$ norm | Capped $L_1$ | ETP | Proposed |
|------|--------|--------|--------|--------|--------|--------|
| 1 | 31.9368 | 31.9668 | 31.2551 | 31.2592 | 32.0769 | **32.4287** |
| 2 | 26.8258 | 26.9897 | 26.0813 | 27.7669 | 27.9989 | **28.0092** |
| 3 | 32.3545 | 32.4669 | 31.4220 | 31.8287 | 32.5875 | **32.9563** |
| 4 | 26.6064 | 26.8509 | 25.9610 | 28.0251 | **28.2247** | 28.1889 |
| 5 | 46.1372 | 46.3494 | 45.2296 | 46.1453 | 46.4733 | **46.9041** |
| 6 | 27.4027 | 27.4787 | 26.8397 | 26.7847 | 27.6708 | **27.9934** |
| 7 | 30.3030 | 30.4297 | 29.4085 | 31.0897 | 31.3823 | **31.5205** |
| 8 | 32.4739 | 32.6596 | 31.5777 | 33.2036 | 33.5479 | **33.6839** |
| Avg. | 31.7550 | 31.8990 | 30.9719 | 32.0129 | 32.4953 | **32.7106** |

| No. | APG | Active ALT | TNNR-ADMM | TNNR-APGL | ETNNR-WRE | TNNR-WRE |
|------|--------|--------|--------|--------|--------|--------|
| 1 | 31.0803 | 31.0786 | 31.9878 | 31.9469 | 32.0243 | 29.3557 |
| 2 | 27.6246 | 27.6242 | 27.5400 | 27.6615 | 27.4193 | 25.3623 |
| 3 | 31.4907 | 31.4886 | 32.7381 | 32.7206 | 32.8937 | 30.2034 |
| 4 | 28.0195 | 28.0193 | 27.6813 | 27.8621 | 27.6011 | 26.3830 |
| 5 | 42.8317 | 42.8330 | 46.2357 | 44.4485 | 45.4205 | 29.4952 |
| 6 | 26.4556 | 26.4539 | 27.3562 | 27.4188 | 27.6793 | 27.0292 |
| 7 | 30.6643 | 30.6628 | 31.0339 | 31.1942 | — | — |
| 8 | 32.8561 | 32.8551 | 33.4827 | 33.5478 | 33.3585 | 30.2777 |
| Avg. | 31.3779 | 31.3769 | 32.2570 | 32.1001 | 32.3424 | 28.3009 |

Natural images usually have low-rank structure. Here, we test on a popular image set that is usually experimented on the image recovery problem [22][23]. The image samples are shown in Fig. 3. The size of image 7 is 105×175 and others have the size of 150×150. We randomly set half of the pixels as corrupted pixels.

The regularizers to compare includes those used in subsection 6.A.1 and some nuclear norm-based methods: APG [29], Active ALT [30], and truncated nuclear norm-based methods: TNNR-ADMM [22], TNNR-APGL [22], ETNNR-WRE [23], and TNNR- WRE [23].

We use the PSNR for recovery quality assessment. We generate 30 different half pixel corruption masks for each image and calculate the average performance. The PSNR comparison is shown in Table IV.

From Table IV, one can see that the proposed method usually has the highest PSNR, which validates the effectiveness of proposed method. Generally, nonconvex regularizers perform better than nuclear norm-based and truncated nuclear norm-based methods. The illustrations of recovery images with proposed regularizer are shown in Fig. 4.

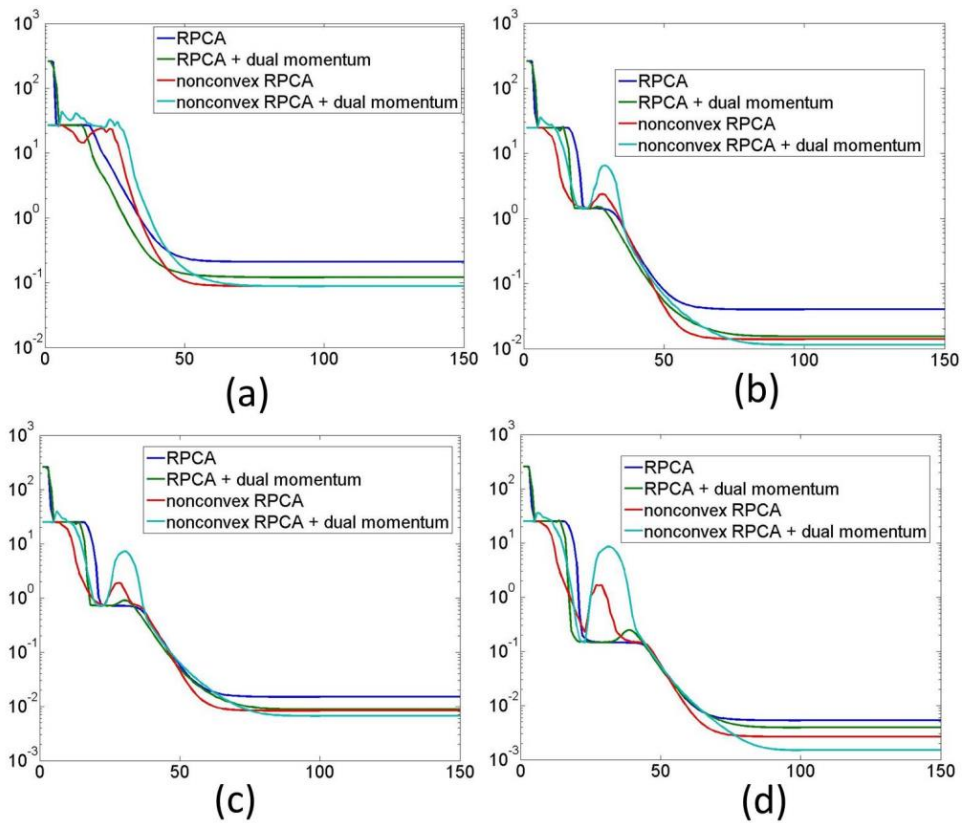

**Fig. 5** Recovered error per iteration for synthetic data experiment in Section VI.B.1. Max iteration is 150. $\alpha = 1$, 0.1, 0.05, and 0.01 for (a), (b), (c), and (d) respectively.

## B. Nonconvex RPCA with Dual Momentum

### B.1 Random matrix synthetic data

First, we generate testing low-rank matrices $M$ and sparse errors $E$ with random supports. Errors with different magnitudes are generated for examination. Testing data matrices $D$ are generated from:

$$D = M + \alpha E \,, \tag{70}$$

where $\alpha$ is a parameter to adjust the error magnitude.

We set $M \in \mathbb{R}^{100 \times 100}$ as a rank-10 matrix whose entries uniformly distribute on [0, 1]. $E$ is a 20% random support error matrix. Four cases $\alpha = 1$, 0.1, 0.05, 0.01 are applied for testing. Based on the success of the proposed nonconvex surrogate for the $rank(.)$, we use the proposed nonconvex regularizer for testing.

To validate the proposed algorithm of nonconvex RPCA with dual momentum, we compare four methods: RPCA with/without dual momentum and nonconvex RPCA with/without dual momentum. For the parameter settings, we set $\lambda = 0.1$, $\mu^0 = 10^{-3}$, $\kappa = 1.2$, and the convergence is achieved when $\left\| D - L^{k+1} - E^{k+1} \right\|_F^2 < 10^{-9}$ or the iteration number $k > 150$.

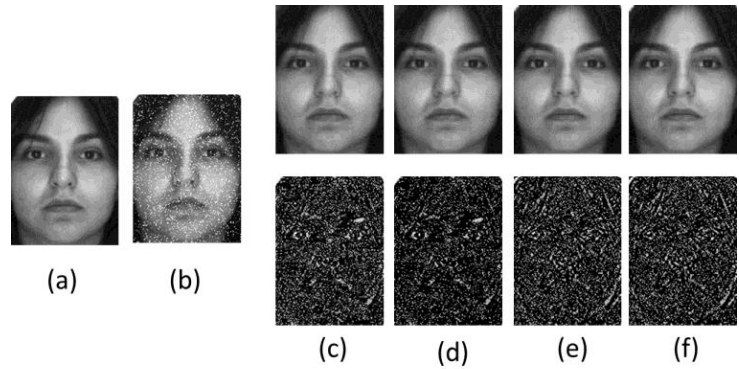

**Fig. 6** Samples of face image denoising. (a) The original image. (b) The corrupted face. (c)(d)(e)(f) The recovered faces (upper row) and recovered noises (lower row) for RPCA, RPCA + dual momentum, nonconvex RPCA, nonconvex RPCA + dual momentum respectively. (For convenience of display, we enhance the magnitude of error in (b) and the lower row of (c)-(f)).

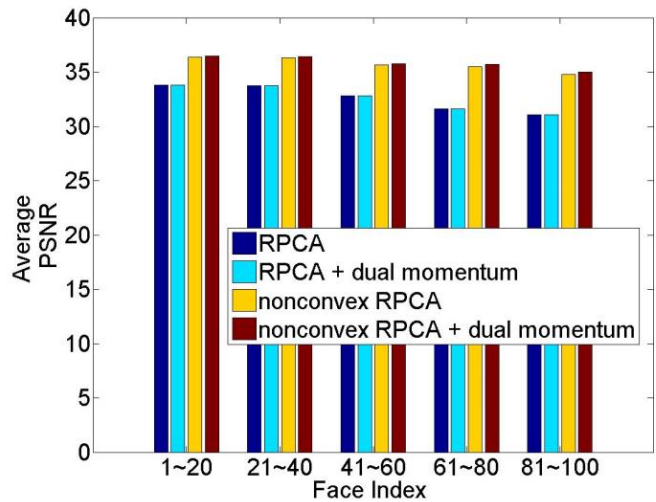

**Fig. 7** Image denoising performance comparison for 100 face images. We calculate the average PSNR every 20 people. Overall, the two proposed methods (nonconvex RPCA and nonconvex RPCA + dual momentum) outperforms the convex RPCA.

TABLE V   The time needed for each method to solve a problem until 100 iterations, averaged for 50 problems. DM is the abbr. of Dual Momentum.

| Time (s) | $\alpha = 1$ | $\alpha = 0.1$ | $\alpha = 0.05$ | $\alpha = 0.01$ |
|---|---|---|---|---|
| RPCA | 0.4900 | 0.4039 | 0.3586 | 0.4526 |
| RPCA+DM | 0.5108 | 0.4195 | 0.3540 | 0.4077 |
| Nonconvex RPCA | 0.4920 | 0.4137 | 0.3511 | 0.4092 |
| Nonconvex RPCA +DM | 0.5159 | 0.4429 | 0.3457 | 0.4115 |

We repeat the tests for 50 times with different randomly generated $E$. We draw the curve of the average of the error, $\| L^k - M \|_F$, per iteration. The result is in Fig. 5.

In the result, we can see that both proposed nonconvex RPCA and dual momentum can effectively enhance the performance, converging to the point with error smaller than RPCA solely. In each case, we can see that nonconvex RPCA with dual momentum leads to the minimal error in every case. RPCA alone converges to the point with maximal error in every case. The curves of RPCA with dual momentum and nonconvex RPCA without dual momentum lie between them.

About time cost, we list the average time to solve a problem until 100 iterations from the fifty problems in Table V. We can see that both introduction of nonconvex regularizer and dual momentum did not bring about additional time cost. Nonconvex RPCA and the convex version have not far-off time cost. This also validates that the proposed methods have linear convergence, as the analysis of Section V.

### B.2 Image denoising on face images

We then test proposed nonconvex RPCA and dual momentum on real-world data.

We use the face images in the *AR* Database [31] for testing. The size of each image is 165×120. We convert the image into grey scale and convert the range of pixel value to [0, 1]. For each of the face, we add the random noise $E$ the same as in Section VI.B.1 with $\alpha = 0.1$. We use the proposed nonconvex function with $(a_1, a_2)$ the same as in Section VI.B.1, and set $(p_1, p_2, p_3) = (1, 40, 60)$. The parameter settings and convergence conditions are the same. Samples of the face data and the recovered image and the error are shown in Fig. 6. Since the real-world image data such as faces is not a strictly low-rank matrix, the residual between the original image and the recovered image are always larger than in the random low-rank matrix case in Section VI.B.1. Therefore, we measure the recovered image quality with PSNR. The result is in Table VI.

From the results in Fig. 7, the proposed nonconvex RPCA outperforms convex RPCA with obvious gap in most cases. Also, nonconvex RPCA with dual momentum has better performance than nonconvex RPCA without dual momentum. This observation coincides with the result in Fig. 5 that in overall, nonconvex RPCA with dual momentum has the best performance. Both the nonconvex approach and the dual momentum are beneficial for the RPCA to converge to a point with smaller recovery residual.

In Table VI, we compare the proposed algorithm to other state-of-the-art RPCA-based methods: STOC-RPCA [38], MoG-RPCA [39], and VBRPCA [40]. One can see that the proposed nonconvex RPCA algorithm with the dual momentum achieves the best average PSNR performance.

TABLE VI   PSNR COMPARISON FOR FACE DENOISING WITH STATE-OF-THE-ART METHODS FROM 100 SAMPLES. DM IS THE

ABBREVIATION OF THE DUAL MOMENTUM.

| PSNR | Average | Min | Standard Deviation |
|---|---|---|---|
| RPCA | 33.5671 | 27.9871 | 3.2775 |
| STOC-RPCA | 30.1352 | 21.0010 | 3.4057 |
| MoG-RPCA | 35.7188 | 27.2506 | 2.5665 |
| VBRPCA | 35.4544 | 32.7549 | **1.2078** |
| Nonconvex RPCA+DM | **37.3041** | **34.0254** | 1.5208 |

Experiments in Section VI.B.1 about synthetic data and Section VI.B.2 about real-world data both validate the efficacies of the proposed nonconvex approach and the dual momentum on RPCA.

## C.  Nonconvex LRR with Dual Momentum

### C.1 Random matrix synthetic data on spectral clustering

In LRR, if the SVD of $Z^*$ is $U^* \Sigma^* (V^*)^T$, then $U^* (U^*)^T$ can recover the row space of the underlying low-rank matrix with guarantee. In [8], the affinity matrix $W$ is constructed as:

$$[W]_{ij} = ([\tilde{U}\tilde{U}^T]_{ij})^2, \quad \tilde{U} = U^* (\Sigma^*)^{(1/2)}. \tag{71}$$

Ideally, the affinity matrix $W$ should be a block diagonal matrix showing the correlations between samples, as in Fig. 8. Then, $W$ is clustered with normalized cut [41] to perform subspace segmentation.

In the experiment, we generate 10 disjoint subspaces. Each of them has the dimension of 10 with ambient dimension 100. We randomly draw 10 samples from each of them and form the testing data matrix $D \in \mathbb{R}^{100 \times 100}$. As in RPCA, we compare the four methods: LRR, LRR with the dual momentum, nonconvex LRR, and nonconvex LRR with the dual momentum. We set $\lambda = 0.1$, $\mu^0 = 10^{-3}$, $\kappa = 1.2$, and the convergence is achieved when $\left\| D - AL^{k+1} - E^{k+1} \right\|_F^2 + \left\| Z^{k+1} - L^{k+1} \right\|_F^2 < 10^{-9}$ or the iteration number $k > 100$. In (33), we set $\gamma = \{2, 1\}$, i.e., the $L_{2,1}$ norm is applied. We use the proposed nonconvex regularizer with $(a_1, a_2) = (0.1, 0.2)$ and $(p_1, p_2, p_3) = (12, 40, 60)$. We repeat the experiments 50 times for different disjoint subspaces and samples and calculate the average clustering accuracy in Table VII.

TABLE VII   CLUSTERING PERFORMANCE COMPARISON OF DIFFERENT METHODS. DM IS THE ABBREVIATION OF THE DUAL MOMENTUM.

| Accuracy | Average | Min | Standard Deviation |
|---|---|---|---|
| LRR | 0.8902 | 0.73 | 0.0687 |
| LRR+DM | 0.8924 | 0.75 | 0.0489 |
| Nonconvex LRR | 0.9134 | 0.71 | 0.0756 |
| Nonconvex LRR +DM | 0.9172 | 0.75 | 0.0682 |

TABLE VIII   ACCURACY COMPARISON ON THE FACE CLUSTERING PROBLEM WITH DIFFERENT METHODS. DM IS THE ABBREVIATION OF DUAL MOMENTUM.

| | LRR | LADM_LRR | WBSLRR |
|---|---|---|---|
| Accuracy | 0.8875 | 0.8953 | 0.9109 |
| | LRR+DM | Nonconvex LRR | Nonconvex LRR+DM |
| Accuracy | 0.8844 | 0.9047 | 0.9141 |

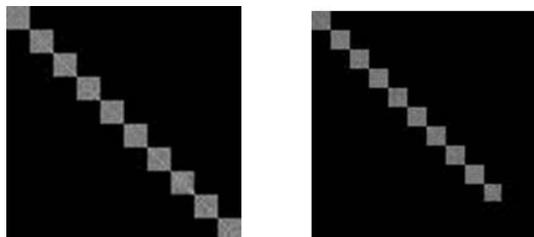

**Fig. 8** (Left) The ideal affinity matrix is block-diagonal. Here, we generate 10 disjoint subspaces and each of them has the dimension of 10. We draw 10 samples from each subspace and after LRR we plot the affinity matrix *W*. (Right) The ideal affinity matrix with outliers. We append 20 uncorrelated outlier samples to the data matrix. The affinities of outliers are 0.

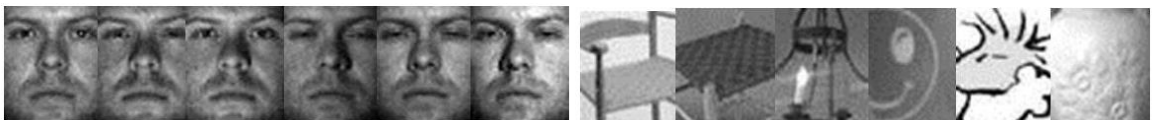

**Fig. 9** Sample images for the combined testing dataset in Section VI.C.2.

In Table VII, one can see that the proposed nonconvex LRR algorithm with the dual momentum has the best average clustering accuracy. The original convex LRR algorithm has the lowest accuracy. This result is similar to that of RPCA. It again validates that both the nonconvex surrogate and the dual momentum can lead to better results than the ordinary ADMM procedure.

*C.2 Spectral clustering anomaly detection on real-world data*

In this experiment, we combine the face data from the Extended Yale Database B [42] and randomly choose 640 face images (64 images with different illuminations of each of the 10 people). Then, we choose 180 object images (9 different objects and 20 images for each) from Caltech101 [43]. Some images for this combined dataset are shown in Fig. 9. The parameter setting is the same as that in Section VI.C.1. We use the proposed nonconvex regularizer with $(a_1, a_2) = (0.1, 0.2)$ and $(p_1, p_2, p_3) = (1, 10, 30)$. For the $K$-means clustering, for evaluation convenience, we remove the outliers from the ground truth as in [8]. We also compare the proposed methods to the advanced methods of WBSLRR [44] and LADM_LRR [45] and. For LADM_LRR, we set $\lambda = 1$ with other parameters unchanged and for WBSLRR we set $\lambda = 0.1$, and $\gamma = 0.5$ to achieve better performances. The accuracy is shown in Table VIII.

From Table VIII, one can see that the proposed nonconvex LRR algorithm with the dual momentum has the best performance. It outperforms the convex LRR, the RR without the dual momentum, WBSLRR, and the LADM_LRR method.

# VII. CONCLUSION

In this paper, a novel nonconvex regularizer is proposed. It can be used as a *rank*(.) surrogate, which has better performance on the rank minimization problem. Then, the proposed nonconvex rank surrogate is applied on two well-known sparse and low-rank models: RPCA and LRR. Moreover, we propose the dual momentum trick to enhance the performance of the ADMM optimization framework which leads to a smaller recovery residual. We also provide complete theoretical convergence analysis on the nonconvex version of the RPCA and LRR with the dual momentum. We prove the boundness of the variables and the KKT optimality of the accumulation points. We also show that the convergence rates for these nonconvex methods are linear, which are the same as those in the convex cases. Extensive experiments are conducted to validate proposed methods, including the image completion problem, denoising on the synthetic and real-world data, and spectral clustering on the synthetic data and real-world data. One can see that the proposed nonconvex regularizer and the proposed nonconvex RPCA and LRR with the dual momentum have even better performances than their convex versions and are useful for these signal and image processing applications.

# APPENDIX

(*Lemma 1*). The following inequality is held for Algorithm 3:

$$\mathcal{L}(L^{k+1}, E^k, \hat{Y}^k, \mu^k) \leq \mathcal{L}(L^k, E^k, \hat{Y}^{k-1}, \mu^{k-1}) + C^k$$

where

$$C^k = \frac{1}{2(\mu^{k-1})^2}\left[(\mu^k+\mu^{k-1})\left\|M^k\right\|_F^2 + 2\frac{\alpha^{k-1}-1}{\alpha^k}\mu^{k-1}\left\|Y^k-Y^{k-1}\right\|_F^2 + 2\frac{\alpha^{k-1}-1}{\alpha^k}\frac{\alpha^{k-2}-1}{\alpha^{k-1}}\mu^{k-1}(Y^{k-1}-Y^{k-2})(Y^k-Y^{k-1})\right],$$

$$M^k = Y^k - (1+\frac{\alpha^{k-2}-1}{\alpha^{k-1}})Y^{k-1} + \frac{\alpha^{k-2}-1}{\alpha^{k-1}}Y^{k-2}.$$

(*Proof*): From (27) we can have

$$\frac{Y^k-\hat{Y}^{k-1}}{\mu^{k-1}} = D - L^k - E^k. \tag{72}$$

From the iteration and (28), we can have

$$\mathcal{L}(L^{k+1},E^k,\hat{Y}^k,\mu^k) \le \mathcal{L}(L^k,E^k,\hat{Y}^k,\mu^k) = \mathcal{L}(L^k,E^k,\hat{Y}^{k-1},\mu^{k-1}) + C^k \tag{73}$$

where

$$C^k = \frac{1}{2}(\mu^k-\mu^{k-1})\left\|D-L^k-E^k\right\|_F^2 + \left\langle \hat{Y}^k-\hat{Y}^{k-1}, D-L^k-E^k \right\rangle = \frac{1}{2}(\mu^k-\mu^{k-1})\left[\left\langle D-L^k-E^k+\frac{2(\hat{Y}^k-\hat{Y}^{k-1})}{\mu^k-\mu^{k-1}}, D-L^k-E^k \right\rangle\right]$$

$$= \frac{1}{2}(\mu^k-\mu^{k-1})\left[\left\langle \frac{Y^k-\hat{Y}^{k-1}}{\mu^{k-1}} + \frac{2}{\mu^k-\mu^{k-1}}\hat{Y}^k - \frac{2}{\mu^k-\mu^{k-1}}\hat{Y}^{k-1}, \frac{Y^k-\hat{Y}^{k-1}}{\mu^{k-1}} \right\rangle\right]. \tag{74}$$

Then, from (27), the first term of the inner product in (74) is:

$$\frac{Y^k-\hat{Y}^{k-1}}{\mu^{k-1}} + \frac{2}{\mu^k-\mu^{k-1}}\hat{Y}^k - \frac{2}{\mu^k-\mu^{k-1}}\hat{Y}^{k-1} = \frac{(\mu^k-\mu^{k-1})(Y^k-\hat{Y}^{k-1}) + 2(\hat{Y}^k-\hat{Y}^{k-1})\mu^k}{\mu^{k-1}(\mu^k-\mu^{k-1})}$$

$$= \frac{(\mu^k-\mu^{k-1})(Y^k-\hat{Y}^{k-1}) + 2\mu^{k-1}\left[(1+\frac{\alpha^{k-1}-1}{\alpha^k})Y^k-\hat{Y}^{k-1}-\frac{\alpha^{k-1}-1}{\alpha^k}Y^{k-1}\right]}{\mu^{k-1}(\mu^k-\mu^{k-1})}$$

$$= \frac{(\mu^k-\mu^{k-1})(Y^k-Y^{k-1}-\frac{\alpha^{k-2}-1}{\alpha^{k-1}}(Y^{k-1}-Y^{k-2}))}{\mu^{k-1}(\mu^k-\mu^{k-1})} + \frac{2\mu^{k-1}\left[(1+\frac{\alpha^{k-1}-1}{\alpha^k})Y^k-Y^{k-1}-\frac{\alpha^{k-2}-1}{\alpha^{k-1}}(Y^{k-1}-Y^{k-2})-\frac{\alpha^{k-1}-1}{\alpha^k}Y^{k-1}\right]}{\mu^{k-1}(\mu^k-\mu^{k-1})}$$

$$= \frac{\left[Y^k-(1+\frac{\alpha^{k-2}-1}{\alpha^{k-1}})Y^{k-1}+(\frac{\alpha^{k-2}-1}{\alpha^{k-1}})Y^{k-2}\right](\mu^k-\mu^{k-1})}{\mu^{k-1}(\mu^k-\mu^{k-1})} + \frac{2\mu^{k-1}\left[(1+\frac{\alpha^{k-1}-1}{\alpha^k})Y^k-(1+\frac{\alpha^{k-1}-1}{\alpha^k}+\frac{\alpha^{k-2}-1}{\alpha^{k-1}})Y^{k-1}+(\frac{\alpha^{k-2}-1}{\alpha^{k-1}})Y^{k-2}\right]}{\mu^{k-1}(\mu^k-\mu^{k-1})} \tag{75}$$

$$= \frac{(\mu^k+\mu^{k-1})\left[Y^k-(1+\frac{\alpha^{k-2}-1}{\alpha^{k-1}})Y^{k-1}+\frac{\alpha^{k-2}-1}{\alpha^{k-1}}Y^{k-2}\right]}{\mu^{k-1}(\mu^k-\mu^{k-1})} + \frac{2\frac{\alpha^{k-1}-1}{\alpha^k}\mu^{k-1}(Y^k-Y^{k-1})}{\mu^{k-1}(\mu^k-\mu^{k-1})}.$$

The second term in the inner product of (74)

$$\frac{Y^k-\hat{Y}^{k-1}}{\mu^{k-1}} = \frac{Y^k-Y^{k-1}-\frac{\alpha^{k-2}-1}{\alpha^{k-1}}(Y^{k-1}-Y^{k-2})}{\mu^{k-1}} = \frac{Y^k-(1+\frac{\alpha^{k-2}-1}{\alpha^{k-1}})Y^{k-1}+\frac{\alpha^{k-2}-1}{\alpha^{k-1}}Y^{k-2}}{\mu^{k-1}}. \tag{76}$$

Combining (75) and (76) to take into (74), we can get

$$C^k = \frac{1}{2(\mu^{k-1})^2}\left[(\mu^k + \mu^{k-1})\left\|M^k\right\|_F^2 + 2\frac{\alpha^{k-1}-1}{\alpha^k}\mu^{k-1}\left\|Y^k - Y^{k-1}\right\|_F^2 - 2\frac{\alpha^{k-1}-1}{\alpha^k}\frac{\alpha^{k-2}-1}{\alpha^{k-1}}\mu^{k-1}\left\langle Y^{k-1} - Y^{k-2}, Y^k - Y^{k-1}\right\rangle\right], \qquad (77)$$

where $M^k$ is the nominator of (76):

$$M^k = Y^k - (1 + \frac{\alpha^{k-2}-1}{\alpha^{k-1}})Y^{k-1} + \frac{\alpha^{k-2}-1}{\alpha^{k-1}}Y^{k-2}. \qquad (78)$$